\documentclass[conference,10pt]{IEEEtran}
\usepackage[nolist]{acronym}
\usepackage[frozencache=true,cachedir=minted_cache]{minted}

\usepackage{tikz}
\usetikzlibrary{external}
\usepackage{graphicx}
\usepackage{amsmath}
\usepackage{listings}
\usepackage{color, colortbl}
\usetikzlibrary{shapes,positioning,automata,arrows,backgrounds,fit}
\def\Act{\operatorname{Act}}

\IEEEoverridecommandlockouts
\usepackage{cite}
\usepackage{amsmath,amssymb,amsfonts}
\usepackage{algorithm}
\usepackage[noend]{algpseudocode} 
\algnewcommand\algorithmicfunctionname{\textbf{function:}}
\algnewcommand\Functionname{\item[\algorithmicfunctionname]}
\algnewcommand\algorithmicinput{\textbf{input:}}
\algnewcommand\Input{\item[\algorithmicinput]}
\algnewcommand\algorithmicoutput{\textbf{output:}}
\algnewcommand\Output{\item[\algorithmicoutput]}
\usepackage{graphicx}
\usepackage{textcomp}
\usepackage{xcolor}
\def\BibTeX{{\rm B\kern-.05em{\sc i\kern-.025em b}\kern-.08em
    T\kern-.1667em\lower.7ex\hbox{E}\kern-.125emX}}

\usepackage{pgf}
\usepackage{afterpage}
\usepackage{hyperref}

\author{\IEEEauthorblockN{Alexander Brauckmann\textsuperscript{\IEEEauthorrefmark{1}}}
\IEEEauthorblockA{\textit{Chair for Compiler Construction} \\
\textit{TU Dresden, Germany} \\
alexander.brauckmann@tu-dresden.de}
\and
\IEEEauthorblockN{Andr\'{e}s Goens\textsuperscript{\IEEEauthorrefmark{1}}}
\IEEEauthorblockA{\textit{Chair for Compiler Construction} \\
\textit{TU Dresden, Germany} \\
andres.goens@tu-dresden.de}
\and
\IEEEauthorblockN{Jeronimo Castrillon}
\IEEEauthorblockA{\textit{Chair for Compiler Construction} \\
\textit{TU Dresden, Germany} \\
jeronimo.castrillon@tu-dresden.de}
}

\usepackage{ifthen}

\usepackage{amssymb}
\usepackage{xcolor}

\newboolean{showcomments}
\setboolean{showcomments}{false}

\makeatletter
\newcommand{\mynote}[3]{   \ifthenelse{\boolean{showcomments}}{    \fbox{\bfseries\sffamily\scriptsize#1}    {\small$\blacktriangleright$\textsf{\emph{\color{#3}{#2}}}$\blacktriangleleft$}}   {    \@bsphack
   \@esphack
  } }
\makeatother

\definecolor{asparagus}{rgb}{0.53, 0.66, 0.42}

\newcommand{\todo}[1]{\mynote{TODO}{#1}{red}}

\begin{acronym}
\acro{SCoP}{Static Control Part} 
\acro{RL}{reinforcement learning} 
\acro{MDP}{markov decision process} 
\acro{IR}{intermediate representation} 
\acro{RNN}{recurrent neural network} 
\acro{GNN}{graph neural network} 
\acro{GCD}{lowest common denominator} 
\acro{LSTM}{long short-term memory} 
\acro{AST}{abstract syntax tree} 
\acro{LCD}{least common denominator} 
\acro{DSL}{domain-specific language} 
\end{acronym}

\begin{document}

\title{A Reinforcement Learning Environment for Polyhedral Optimizations}

\maketitle

\begingroup\renewcommand\thefootnote{\IEEEauthorrefmark{1}}
\footnotetext{Equal contribution}
\endgroup

\begin{abstract}
	The polyhedral model allows a structured way of defining semantics-preserving transformations to improve the performance of a large class of loops.
Finding profitable points in this space is a hard problem which is usually approached by heuristics that generalize from domain-expert knowledge.
Existing problem formulations in state-of-the-art heuristics depend on the shape of particular loops, making it hard to leverage generic and more powerful optimization techniques from the machine learning domain.
In this paper, we propose PolyGym, a shape-agnostic formulation for the space of legal transformations in the polyhedral model as a Markov Decision Process (MDP).
Instead of using transformations, the formulation is based on an abstract space of possible schedules.
In this formulation, \emph{states} model partial schedules, which are constructed by \emph{actions} that are reusable across different loops.
With a simple heuristic to traverse the space, we demonstrate that our formulation is powerful enough to match and outperform state-of-the-art heuristics.
On the Polybench benchmark suite, we found transformations that led to a speedup of 3.39x over LLVM O3, which is 1.83x better than the speedup achieved by ISL.
Our generic MDP formulation enables using reinforcement learning to learn optimization policies over a wide range of loops.
This also contributes to the emerging field of machine learning in compilers, as it exposes a novel problem formulation that can push the limits of existing methods.

\end{abstract}


\section{Introduction}
In most compute-intensive applications, a significant amount of the execution time is spent in loops.
Loop optimization has thus received plenty of attention in the compiler community, where one finds as diverse strategies for optimization as are the loops themselves.
Among them, polyhedral compilation is an important class of strategies for loop optimization which focuses on a particular class of (nested) loops with regular properties~\cite{graphite,isl,grosser2011polly}.
While this significantly restricts the family of loops that can be optimized, polyhedral compilation methods are very relevant in practice.
Compilers for important domain-specific languages like TensorComprehensions~\cite{tensorcomprehensions} and Tiramisu~\cite{baghdadi2019cgo} build atop these methods.
Polyhedral compilation methods have also been used for efforts on recognizing \emph{motifs}, or well-known high-level computations~\cite{gareev2018high}.
The models have been extended beyond explicit loops to support recursive functions calls~\cite{kobeissi2020rec2poly}, or used to target emerging memory technologies by exploiting the detailed information in the model~\cite{vadivel2020date,khan_cases20}.

With the polyhedral model we can find semantics-preserving transformations that significantly improve the loops' performance. 
We do this by exploring constraints exposed by geometric properties of the dependencies between statements in different iterations of the loops. 
However, this exploration requires considering a prohibitively large space of possible code transformations.
The two main methods to explore this space are heuristics, which use domain-specific models to directly find a good candidate transformation, and iterative meta-heuristics, which iteratively explore the space by evaluating multiple points and adapting the solutions.
Model-based heuristics, like Pluto~\cite{pluto} and ISL~\cite{isl}, are more applicable in practice, but the resulting code is consistently outperformed by significantly more time-costly iterative meta-heuristics, like~\cite{pouchet2008iterative,ganser2017iterative}.

A promising strategy to bridge this gap is \acf{RL}.  \ac{RL} has proven to be very successful for navigating vast discrete spaces, while training a decision policy that can later be leveraged to find good solutions directly~\cite{mnih2015human,silver2016mastering}.
Precisely for this reason, \ac{RL} is well-suited for learning heuristics in the spaces of compiler optimizations, seeking a sweet spot between applicable heuristics and iterative methods:
While iterating the search space, a heuristic model is learned that can later be re-used in future iterations to lead to a profitable transformation.

No \ac{RL} method has been applied to the search spaces of polyhedral loop scheduling yet. \todo{Add examples} This is partly because the common way of thinking of loop optimizations in terms of transformations, like  loop fusion, fission  or tiling, is not well-suited for \ac{RL}.
The first disadvantage concerns correctness: not every transformation can be applied to every schedule.  This happens often and is computationally costly.
The greater problem is that the transformations themselves depend on the concrete instance of the problem.
For example, loop fission needs to select the concrete statements and loop bounds, which depend on the concrete loop.
If we encode these transformations as actions in a \ac{MDP} for \ac{RL}, any policy for finding profitable schedules would be specific to that loop or kernel by the very structure of the formulation.

In this paper we propose a formulation that overcomes the limitations of transformations for an \ac{MDP}.
We utilize a known formulation~\cite{feautrier1992schedulingI,feautrier1992schedulingII} that leverages Farkas' lemma to construct an abstract space of exclusively valid schedules.
We give present a clear overview of the background from discrete geometry, emphasizing why the different steps in the construction are important for this use-case (Section~\ref{sec:poly}).
 Our \ac{MDP} navigates this abstract space in an instance-independent fashion by leveraging its algebraic properties (Section~\ref{sec:mdp}).
The resulting search space boasts the potential for achieving great performance and learning, a fact we demonstrate by traversing it with simple heuristics (Section~\ref{sec:eval}).
Even without sophisticated \ac{RL} algorithms, simple heuristics that guide the iterative process outperform the state-of-the-art heuristics in ISL~\cite{isl} in the Polybench benchmark suite~\cite{polybench} in a small number of iterations.     
 Importantly, as well, the field of machine learning for compilers is an emerging field of study on its own~\cite{wang2018machine}.
\ac{RL} algorithms are a particularly relevant class of methods for compiler optimization~\cite{leather2020machine}, and have successfully been applied to certain subproblems already~\cite{haj2020neurovectorizer,CompilerGym,mammadli2020static}.
Even in the industrial LLVM compiler, a \ac{RL}-trained heuristic for inlining was recently integrated into the mainline version~\cite{trofin2021mlgo}.\footnote{\url{https://github.com/llvm/llvm-project/blob/main/llvm/lib/Analysis/InlineAdvisor.cpp}} However, defining \ac{RL} environments for problems in compiler optimization is challenging in general, for many of the same reasons it is challenging in the case of polyhedral optimizations.
Defining the problem in a way that action policies can be learned in a uniform fashion, generalizing across multiple samples, is an important step to advance the field of machine learning in compilers as well.

\section{Background: Polyhedral Compilation}
\label{sec:poly}

\begin{figure}[h]
	\centering
  \resizebox{0.25\textwidth}{!}{
    \includegraphics{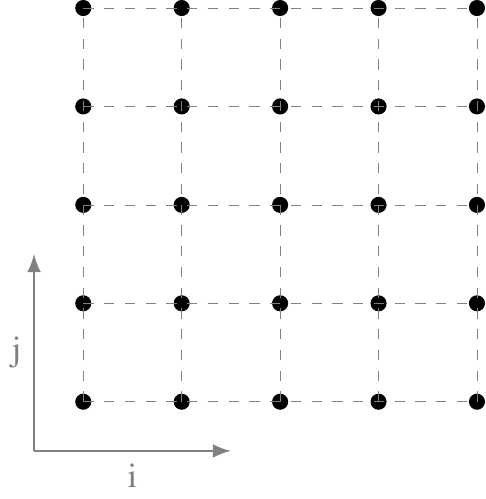}
  }
   \caption{The lattice polytope interpretation of the iteration space of the matrix-vector multiplication kernel for \texttt{N=5}.}
	\label{fig:iteration_space_matvec}
\end{figure}

This section introduces the basics of the polyhedral model, based on the original work of Feautrier~\cite{feautrier1992schedulingI,feautrier1992schedulingII}, and the algorithms for schedule space construction due to
Pouchet and others~\cite{pouchet2007iterative,pouchet2008iterative}.
Consider the following C kernel for the matrix-vector multiplication $y = Ax$:

\begin{minted}{C}
      int i, j;
      for (i = 0; i < N; i++) 
S:     y[i] = 0;
      for (i = 0; i < N; i++) 
        for (j = 0; j < N; j++)
T:       y[i] += A[i][j]*x[j];
\end{minted}

The iterations of the loops in this kernel exhibit a regular structure.
The set of points $i,j$ of the values the variables \texttt{i,j} take during the execution of the loops is $I := \{ i, j \in \mathbb{Z} \mid 0 \leq i,j \leq N-1 \}$.
This kernel is what we call a \acf{SCoP}.
A detailed description of \acp{SCoP} and how they can be constructed can be found in~\cite{grosser2011polly}.
For the purposes of this paper however, it suffices to know that these programs have regular iteration spaces.
Concretely, if we consider the iteration variables in the loops of the \ac{SCoP}, they have a geometric interpretation as a lattice polytope.
A polytope is the $n$-dimensional generalization of a polygon in $2$ dimensions and polyhedral in $3$ dimensions.
The lattice contains a set of discrete points (e.g. integer values) in the interior of the polytope.
For example, the set $I$ of the example kernel can be seen as the discrete points in the lattice $\mathbb{Z}^2$, restricted to an $N \times N$ square starting at the origin, as depicted in Figure~\ref{fig:iteration_space_matvec}.

\begin{figure}[h]
	\centering
  \resizebox{0.45\textwidth}{!}{
    \includegraphics{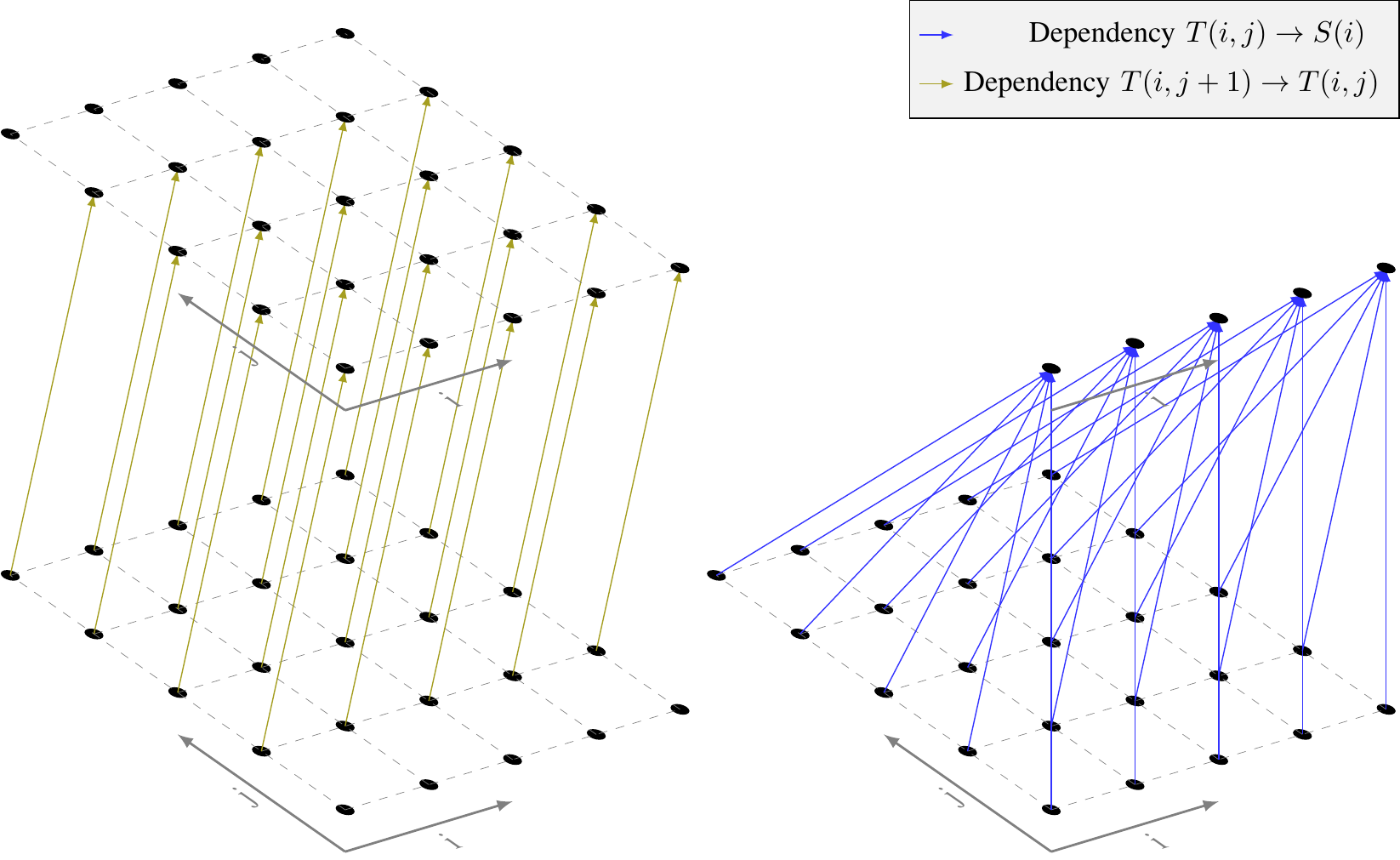}
  }
   \caption{The geometric interpretation of dependencies in the matrix-vector multiplication kernel for \texttt{N=5}.}
	\label{fig:dependencies_matvec}
\end{figure}

In the kernel, there are two distinct statements that are executed in the loop, which we labelled by $S$ and $T$.
The first statement $S :$ \texttt{y[i] = 0} does not actually depend on $j$, and is executed $N$ times, which we denote by $S(i)$ for $i = 0,\ldots,N-1$.
The statement $T :$ \texttt{y[i] += A[i,j]*x[j]} depends on both variables $i,j$ and is executed $N^2$ times, which we denote by $T(i,j)$ for $i,j = 0,\ldots,N-1$.
In particular, executing this kernel amounts to executing these $N^2+N$ statements.
Analyzing the statements we can see which statements have to be executed before which, and where it does not matter.
Indeed, we know that for every $i$, the instance $S(i)$ has to be executed before all $T(i,j)$. 
Similarly, we need to execute $T(i,j)$ before we execute $T(i,j+1)$ for all valid $i,j$.
\footnote{If we know that addition is commutative, we can relax these conditions on the $T(i,j)$, but we ignore the commutativity of addition for this discussion.}
We can use the geometric interpretation as polytopes (in this example, polygons) to visualize these dependencies, as in Figure~\ref{fig:dependencies_matvec}.
  
Any permutation of the $N^2 + N$ statements $S(i), T(i,j)$ that respects the dependencies visualized in Figure~\ref{fig:dependencies_matvec} will produce the same (correct) result.
They are all semantics-preserving permutations.
For example, the original code executes the statements in the order $S(0), S(1), \ldots, S(N-1), T(0,0), T(0,1), \ldots, T(N-1,N-1)$. 
However, we could also execute them in the order $S(0), T(0,0), S(1), T(0,1), T(1,0), S(2), \ldots, S(N-1), T(0,N-1), \ldots, T(N-1,0), T(1,N-1),\ldots,T(N-1,N-2), T(N-1,N-1)$, as this respects all dependencies.
This order would translate to the code: 
\begin{minted}{C}
int i, j;
for (i = 0; i < N; i++){
  y[i] = 0;
  for (j = 0; j <= i; j++)
    y[j] += A[j][i-j]*x[i-j];}
for (i = 0; i < N; i++)
  for (j = i+1; j < N; j++)
    y[j] += A[j][i-j+N]*x[i-j+N];
\end{minted}

The semantics-preserving permutations of the statements can be represented as \emph{schedules}.
Using the lexicographic ordering, we can encode schedules as vectors. 
For example, the original statement order for $N=5$ could be encoded as the $1$-dimensional vectors for $S(i) = (i), \text{ and } T(i,j) = i+5 \cdot j+5$ respectively, which results in the order $(0),\ldots,(4)$ for $S$ and $(5),\ldots,(29)$ for $T$, encoding the order of the statement executions.
The crucial observation is that these vectors are affine functions
\begin{align*}
 & \Theta_S: \mathbb{Z}^2 \rightarrow \mathbb{Z}^1, (i,j) \mapsto (1,0,0)^T(i,j,1), \\
 & \Theta_T: \mathbb{Z}^2 \rightarrow \mathbb{Z}^1, (i,j) \mapsto (1,5,5)^T(i,j,1).
\end{align*}
This schedule is called the \emph{identity schedule}, as it represents the original untransformed code.

In general, when the schedule transformations are affine linear functions, we can use a powerful result of discrete geometry called Farkas' lemma in its affine form~\cite{schrijver} to characterize all such transformations~\cite{feautrier1992schedulingI}.
The transformed schedule we showed above, for example, has the order $0,2,5,9,14$ for the instances $S(i), i= 1, \ldots, 4$, which is not an affine function of $i$.
We can express the schedule as an affine function, however, if we use two-dimensional vectors $S(i) = (i,0)$ and $T(i,j) = (i+j,i+1)$.
These vectors, sorted lexicographically, give the order of the schedule above and are affine functions as well. 
When scheduling \acp{SCoP}, this characterization results in a set of linear inequalities describing the valid schedules~\cite{pouchet2007iterative}.
This set of inequalities comes about by encoding each dependency between statements as a precedence constraint on the iteration spaces.
Farkas' lemma then allows us to transform affine functions respecting these dependencies into this set of inequalities.
For example, for the matrix-vector multiplication kernel, the valid 1-dimensional schedules are integer solutions to the inequalities on the $11$ variables $i_0, \ldots, i_{10}$
\begin{equation}
  \begin{array}{llllll}
&  i_2 > 0 & \text{and} & i_6 \geq i_3   & \text{and} & i_7 \geq i_4 \\
\text{and} & i_8 \geq i_5 &  \text{and} & \multicolumn{3}{l}{i_7 \geq i_0 - i_1 + i_3 + i_4 - i_6} \\
\text{and} & \multicolumn{5}{l}{  i_{10} > i_3 + i_4 + i_5 - i_6 - i_7 - i_8 + i_9.}\\
  \end{array}
  \label{eqn:constraints}
\end{equation}

\begin{figure}[h]
	\centering

  \resizebox{0.45\textwidth}{!}{
    \includegraphics{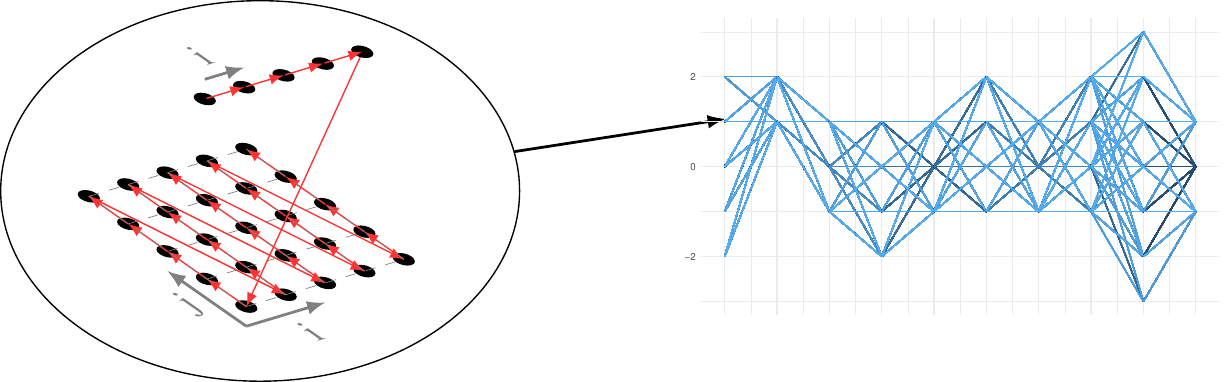}
  }
  \caption{The geometric interpretation of the space of valid one-dimensional schedules in the matrix-vector multiplication kernel for \texttt{N=5}.
  The space is a lattice polytope on $11$ dimensions, visualized as a parallel plot.}
	\label{fig:schedule_space_matvec}
\end{figure}

Geometrically, this set of inequalities can be equivalently considered as another lattice polytope, where each point in this polytope represents a valid schedule.
Importantly, this polytope should not be confused with those for the iteration space or the dependencies, where the points represent the values of the iteration variables.
Figure~\ref{fig:schedule_space_matvec} depicts this set for one-dimensional schedules. The space of valid schedules is itself a polytope in $11$ dimensions, corresponding to the $11$ variables in Equation~\ref{eqn:constraints}.
  As such, we depict it with a parallel plot, where each horizontal line in the grid is a dimension and each point has a different color.
In fact, this lattice polytope has an infinite number of points; we show only the meet of the polytope with the fundamental region of the lattice $(\cong \mathbb{Z}^{11})$ in this plot.
Each point represents a schedule, like the original schedule which is depicted explicitly in the figure.
Note that one-dimensional refers to the schedule vectors, i.e. the images of $\Theta_S : \mathbb{Z} \rightarrow \mathbb{Z}^1$ and $\Theta_T : \mathbb{Z}^2 \rightarrow \mathbb{Z}^1$.
This dimension is different from the points depicted, which encode the constraints as characterized by Farkas's lemma.
The polytope corresponding to the constraints in Equation~\ref{eqn:constraints} includes both schedule functions, $\Theta_S$ and $\Theta_T$.

Any valid schedule can be expressed as a one-dimensional transformation function. The caveat is, however, that this transformation is not necessarily an affine function.
As such, Farkas' lemma does not apply, and we cannot capture these schedules in the lattice polytope of valid one-dimensional transformations.
We can use more dimensions to describe more schedules as affine functions~\cite{feautrier1992schedulingII}.
To define a $k$-dimensional schedule, we define $k$ different polytopes $P_1, \ldots, P_k$ using Farkas' lemma, as described above for the one-dimensional case (cf. Figure~\ref{fig:schedule_space_matvec}).
Each polytope $P_d$ encodes the functions for the corresponding component, i.e. $(\Theta_S)_d, (\Theta_T)_d$.
In the example of the two-dimensional schedule for the matrix-vector multiplication example, we have $(\Theta_S)_1(i,j) = i, (\Theta_S)_2(i,j) = 0$ and similarly, $(\Theta_T)_1(i,j) = i+1, (\Theta_T)_2(i,j) = i+1$.
Each of these is an affine function of the inputs.

The characerization of the space of \emph{all} possible multi-dimensional schedules is not as straightforward as it is in the one-dimensional case.
For this we distinguish between two types of precedence constraints. 
For a dependency $S \rightarrow T$ we say that the polytope $P_i$ carries the dependency \emph{strongly} if it encodes that the value of $\Theta_S < \Theta_T$.
If instead it only encodes the weaker constraint  $\Theta_S \leq \Theta_T$, we say that $P_i$ carries this dependency \emph{weakly}.
While this does not enforce the constraint, it ensures that this constraint is not violated either.
 A proper definition of these concepts can be found in~\cite{pouchet2008iterative}.
For an example, consider the dependency $S \rightarrow T$ in the matrix vector multiplication.
In the the two-dimensional schedule defined above, this dependency is carried weakly by the polytope $P_1$ corresponing to the first dimension, since $i \leq i + j$ but $i \nless i + j$ for $(i,j) \in I$. The dependency $T \rightarrow T$, on the other hand, is carried strongly by $P_1$, since $i + j < i + (j+1)$ for all $(i,j) \in I$.
The polytope $P_2$ for second dimension then carries the dependency $S \rightarrow T$ strongly, since $0 < i + 1$ for all $(i,j) \in I$.
 
If we consider a polytope carrying all dimensions strongly, we leave out some possible schedules.
In some cases, the polytope carrying all dimensions strongly might even be empty, an example of this is given in~\cite{feautrier1992schedulingII}.
Instead, we can iteratively construct a multi-dimensional schedule by carrying only some dimensions strongly and ensuring all the other dependencies are carried weakly.
Since we interpret schedules in lexicographical order, once a dependency is carried strongly by a dimension, all subsequent dimensions need not carry it weakly anymore.
If we iteratively add dependencies to dimensions in this way, we will always find a multi-dimensional schedule for every \ac{SCoP}~\cite{feautrier1992schedulingII}.
However, the order in which we let dimensions carry the dependencies strongly affects the space of possible schedules.
In Section~\ref{sec:mdp} we define a general algorithm with free parameters to explore all possible such constructions.

To find profitable schedules for a \ac{SCoP}, we need to choose a schedule after having constructed the space of all possible (multi-dimensional) schedules.
For this, we can independently consider the polytopes $P_d$ corresponding to each dimension $d$ of the schedule space.
Points in this schedule space are integer solutions to a set of inequalities, but equivalently, they are points with integer values in a special linear combination of terms:
\begin{align}\label{eq:terms} p = \sum_i^{s} \lambda_i v_{i,d} + \sum_i^{t} \alpha_i r_{i,d}, \lambda_i, \alpha_i \in \mathbb{R}_{\geq 0}, \sum_i^{k_1} \alpha_i = 1. \end{align}
Equation~\ref{eq:terms} describes a convex combination of terms $v_{i,d}$ called vertices, and a positive linear combination of terms $r_{i,d}$ called rays, where the index $d$ corresponds to the dimension $d$.
Every such combination with integer values, i.e., where $p \in \mathbb{Z}^n$, is a point in the lattice polytope of valid schedules.
In other words,
\[P_d = \mathbb{Z}^n \cap \{\sum_i \lambda_{i} v_{d,i} + \sum_i \alpha_{i} r_{d,i} \mid \lambda_i, \alpha_i \geq 0, \sum_i \lambda_i = 1 \}. \]
This representation can be calculated with Chernikova's algorithm~\cite{chernikova} and is equivalent to the constraint set representation (as e.g. in Equation~\ref{eqn:constraints}).
We call the set $G_d = \{v_{d,1},v_{d,1},\ldots,r_{d,1},\ldots \}$ the set of \emph{generators} of $P_d$.
 \todo{how does this change with the multiple dimensions?}
  Choosing a point $p_d \in P_d$ in the corresponding polytope for every dimension thus defines a multi-dimensional schedule. 

\begin{figure}[h]
	\centering
  \resizebox{0.45\textwidth}{!}{
    \includegraphics{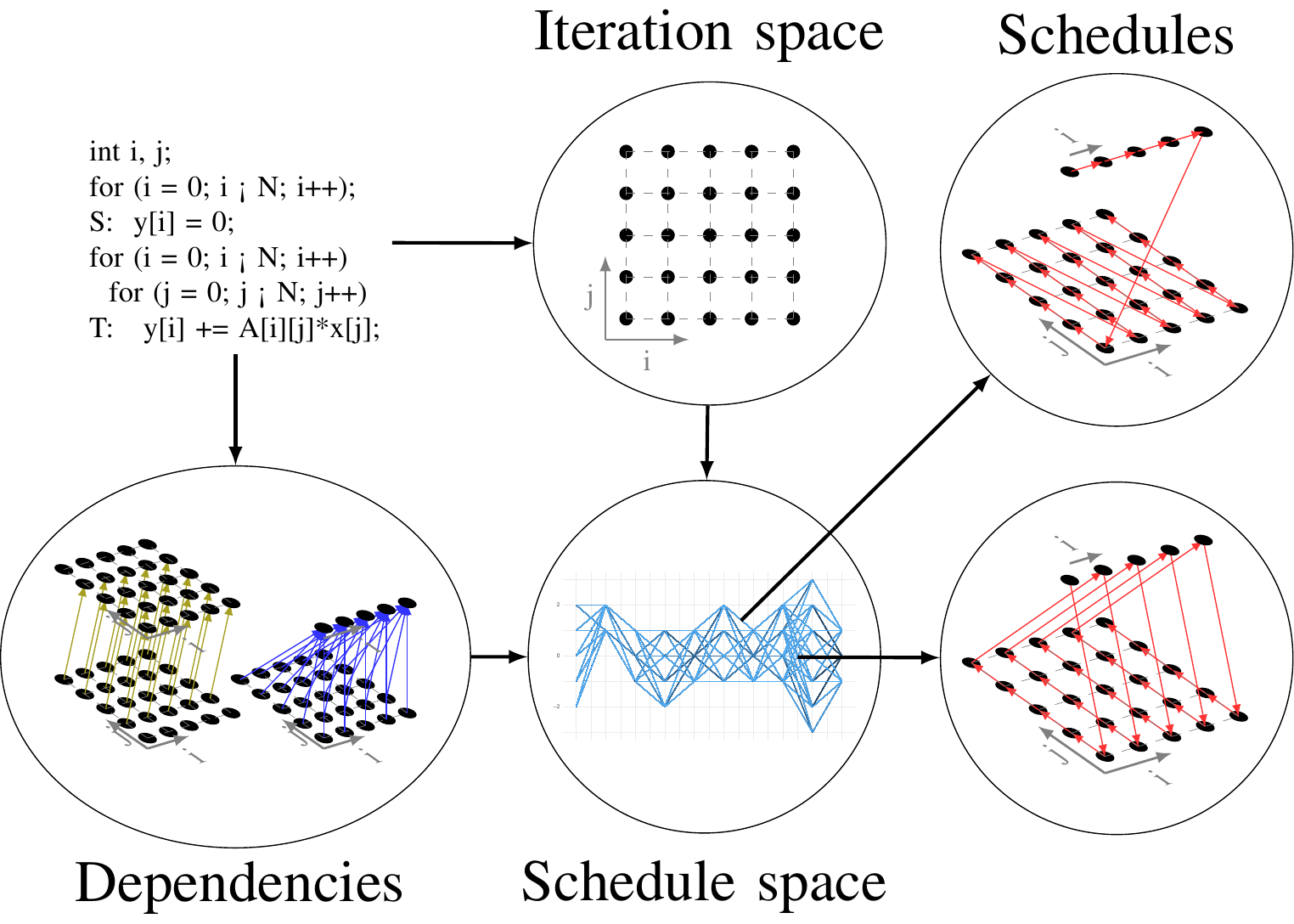}
  }
	\caption{An overview of schedules in the polytope model}
	\label{fig:poly_overview}
\end{figure}

As we have seen, a confusing aspect of the polyhedral model is how there are several lattice polytopes associated to a \ac{SCoP} region: the iteration space, geometric interpretations of the dependencies, and the spaces of valid schedules.
Figure~\ref{fig:poly_overview} gives an overview of the model and the different lattice polytopes, as well as the relationships between them, on the basis of the matrix-vector multiplication example.

Optimizing a loop amounts to finding a good schedule for the loop and a target architecture.
Since the construction of a multi-dimensional space of valid schedules is not unique, this can be considered as two distinct steps:
Constructing a schedule space (of valid schedules) and exploring it to find a profitable schedule.
An important contribution of this paper are \ac{MDP} models for finding such a schedule based on these two steps, as we describe in Section~\ref{sec:mdp}.
In Section~\ref{sec:related} we describe how state-of-the-art methods go about finding them.

\section{Related Work}
\label{sec:related}
There are two main families of methods for schedule optimization in the polyhedral model: iterative methods that are often guided by meta-heuristics and model-driven heuristics.
Iterative models evaluate multiple schedules iteratively by generating code and benchmarking it, eventually returning the best schedule~\cite{pouchet2007iterative}.
The iterative exploration of such large spaces is typically guided by meta heuristics such as genetic algorithms, where the results of evaluations influence future sampling points in the space~\cite{pouchet2008iterative,ganser2017iterative,ganser2018speeding}.
Since these methods require multiple executions of the program, they are time-consuming in practical compilation flows and need additional benchmark tooling.
In contrast to this, model-driven heuristics like Pluto~\cite{pluto} and ISL~\cite{isl} propose a reasonably good solution directly without benchmarking.
Using predefined assumptions about communication and locality, these approaches formulate the optimization problem as an Integer Linear Program. 
The assumptions require domain knowledge that may not generalize to other architectures, and are usually outperformed by their iterative counterparts. 
Learned heuristics for schedule optimization have been investigated in the scope of \acp{DSL} that do loop-heavy computations, like TensorComprehensions for machine learning~\cite{tensorcomprehensions} or Halide~\cite{halide} for image processing.
A recent extension to Halide~\cite{adams2019learning} uses machine learning on a set of program features to build a cost model. 
The authors use the cost model to steer a beam search to find profitable schedules iteratively without benchmarking the programs.
In similar fashion, TVM~\cite{chen2018tvm} uses a learned cost model to estimate the performance of a candidate schedule in its form of a program representation.
TensorComprehensions, Halide and TVM work on a similar space of loop transformation primitives.
Compared to the search space presented in this work, their space does not include affine transformations that lead to potentially fast loop variants. These methods are also restricted to programs expressed in the corresponding \acp{DSL}.
In contrast, this work presents a search space that is applicable to general C code.

Recent work using Tiramisu~\cite{baghdadi2021deep} overcomes some of these limitations by considering polyhedral optimizations as well.
In addition to that, they use machine learning to estimate the execution time of a loop after the transformation, but use an iterative monte-carlo tree search method to explore the possible transformations.
Our contribution is orthogonal to their approach.
Unlike our formulation, however, the search-space in~\cite{baghdadi2021deep} includes incorrect transformations and, more importantly, is specific to the loop being optimized.
This makes it more challenging to design and learn a policy that generalizes across different loops.

\ac{RL} has successfully been applied to certain subproblems of compiler optimizations.
NeuroVectorizer learns a model that aims to predict a solution in the search space of vectorization factors for a given loop~\cite{haj2020neurovectorizer}.
Other work has focused on maximizing speedup~\cite{mammadli2020static} and code size~\cite{CompilerGym} on the search space of analysis and transformation passes in LLVM, which is enormously large. 
A smaller search space for code size optimization is used by~\cite{trofin2021mlgo} to predict whether or not to inline a function.
In PolyGym, we construct a medium-sized search space that contains many profitable solutions, as we will show in Section~\ref{sec:eval}.
  
More generally, machine learning has recently been applied successfully to multiple tasks in compilers~\cite{cummins2017end,brauckmann_cc20,ye2020deep,cummins2020programl}.
While these works have advanced the models of code, they have focused on rather coarse-grained tasks, like mapping OpenCL kernels to a CPU or GPU.
An important reason for this is that formulating problems in compilers in an instance-independent fashion is challenging.
Finding optimal loop schedules is a more subtle problem than, for instance, selecting a particular GPU or CPU for the execution of a kernel. 
This paper contributes to the field of machine learning in compilers by providing more challenging, yet lucrative task that can better challenge the models of code.

\section{PolyGym's Markov Decision Process}
\label{sec:mdp}

In this section we present our formulation of the problem of finding a profitable schedule as a \acf{MDP}.
This enables us to explore the space of possible schedules for a \ac{SCoP} in the context of \ac{RL}.
In particular, the actions in our formulation are independent of the particular instance, i.e. the \ac{SCoP} being optimized.
If, instead, the actions depended on the particular instance, then a different model needs to be used for every instance.
This property is thus a necessary requirement for a heuristic to learn to generalize over different instances.

As explained in Section~\ref{sec:poly}, the problem of finding a profitable schedule can be divided in two stages: the construction of the schedule space and the selection of a schedule.
We define two different \acp{MDP}, one for each subproblem.
From the point of view of the formulation, these are two distinct \acp{MDP}, even if the space of the state space of the second \ac{MDP} depends on the options chosen in the first.
Thus, a single combined \ac{RL} agent can operate on the combined space, or alternatively, two distinct agents can be trained to explore them.

\subsection{Schedule space construction}
We look for a schedule space in its general form, and thus define it as a multi-dimensional space~\cite{feautrier1992schedulingII,pouchet2008iterative}.
As illustrated in Algorithm~\ref{algo:construction}, we construct a $k$-dimensional schedule space iteratively, by going through the dimensions of the space.
For each dimension, we decide which dependencies to include as strong dependencies, until we have selected all dimensions.
Crucially, the function that decides this, \texttt{select\_dependency}, is left as an unspecified, free function.
It then calculates the polytope of possible schedules strongly satisfying these dependencies and weakly satisfying the remaining unselected dependencies.
Using Chernikova's algorithm~\cite{chernikova} it calculates generators as vertices and rays for this polytope.
This algorithm goes back to the principles outlined by Feautrier~\cite{feautrier1992schedulingII}, which is the same basis for the heuristic in Section 3.2 in~\cite{pouchet2008iterative} and Algorithm~1 in~\cite{ganser2017iterative}.
Algorithm~\ref{algo:construction} generalizes these principles by leaving the decision function \texttt{select\_dependency} unspecified, instead of proposing a concrete heuristic.
Note that we write \texttt{select\_dependency}($d$) to specify that this function depends on the representation of the dependency $d$. 
In general, this decision could also depend on other parameters like properties of the \ac{SCoP}.

\begin{algorithm}
	\caption{General construction of the schedule space}
	\label{algo:construction}
	\begin{algorithmic}[1]
	  \Input{A set $D$ of dependencies}
	  \Output{A schedule space $S = (G_1,G_2,\ldots,G_k)$, where each $G_i$ corresponds to the set of generators of the lattice polytope $P_i$for the $i$-th dimension.}
    \State $i \leftarrow 1$
    \While{ $D \neq \emptyset$}
    \State $\operatorname{Deps}_i \leftarrow \emptyset$
	  \For{ $d \in D$}
    \If{ \texttt{select\_dependency(d)}}
    \State $\operatorname{Deps}_i \leftarrow \operatorname{Deps}_i \cup \{ d \}$
    \EndIf
    \EndFor
    \State $D_i \leftarrow \texttt{strong\_deps}(\operatorname{Deps}_i) \cap \texttt{weak\_deps}(D \setminus \operatorname{Deps}_i)$
    \State $G_i \leftarrow \texttt{chernikova}(D_i)$
    \State $i \leftarrow i + 1$
    \EndWhile
	  \Return $(G_1,\ldots,G_{i-1})$
\end{algorithmic}
\end{algorithm}

This simple change, making \texttt{select\_dependency} a free function, has profound consequences.
The greedy heuristics of~\cite{feautrier1992schedulingII,pouchet2008iterative} construct one deterministic schedule space, which fundamentally limits the space of possible schedules that can be found with the method.
On the other hand, the randomized heuristic of~\cite{ganser2017iterative} could in principle produce any schedule space.
The statistical bias of the randomized algorithm, however, implies that this is never the case in practice, by the law of big numbers.
While this is good for finding good heuristics in most cases, it fundamentally limits the ceiling of possible improvement.
By leaving \texttt{select\_dependency} unspecified, a model could \emph{learn} a good heuristic, which can also leverage properties of the concrete instance of the problem.

Based on Algorithm~\ref{algo:construction} we can define an \ac{MDP} that constructs this space.
  It considers the free \texttt{select\_dependency} function as an action, which is combined with two additional actions for controlling the iteration in from Algorithm~\ref{algo:construction}.
This allows a walk through the \ac{MDP} to steer the iteration through the construction.
     The state space of the \ac{MDP} is the countably infinite space
\begin{align*} S_\text{cons} = \{ (i_\text{dim},i_\text{dep},d_1,\ldots,d_{|D|}) \\ \mid i_\text{dim},i_\text{dep},  d_1,\ldots,d_{|d|}  \in \mathbb{N}, i_\text{dim} > 0 \}. \end{align*}
In this space, the first component $i_\text{dim}$ represents the dimension, the second $i_\text{dep}$ represents the current dependency being selected, while the other components represent the strong dependencies included in that dimension. We define the set of actions as $\Act_\text{cons} = \{ \texttt{next\_dim}, \texttt{next\_dep}, \texttt{select\_dep}\}$.
The transition probabilities $\textbf{P}_\text{cons} : S_\text{cons} \times \Act_\text{cons} \times S_\text{cons} \rightarrow [0,1]$ are defined as follows:
\begin{align*}
  \textbf{P}_\text{cons}((i_\text{dim},i_\text{dep},d_1,\ldots,d_{|D|}),\texttt{next\_dim}, \\
  (i'_\text{dim}+1,i'_\text{dep},d'_1,\ldots,d'_{|D|}) ) \\
   = \delta_{i_\text{dim},i'_\text{dim}} \cdot \delta_{i_\text{dep},i'_\text{dep}} \cdot \delta_{d_1,d'_1} \cdots \delta_{d_{|D|},d'_{|D|}}, \\
  \textbf{P}_\text{cons}((i_\text{dim},i_\text{dep},d_1,\ldots,d_{|D|}),\texttt{next\_dep},\\
  (i'_\text{dim},i'_\text{dep}+n,d'_1,\ldots,d'_{|D|}) ) \\
   = \delta_{i_\text{dim},i'_\text{dim}} \cdot \delta_{i_\text{dep},i'_\text{dep}} \cdot \delta_{d_1,d'_1} \cdots \delta_{d_{|D|},d'_{|D|}}, \\
  \textbf{P}_\text{cons}((i_\text{dim},i_\text{dep},d_1,\ldots,d_{|D|}),\texttt{select\_dep},\\
  (i'_\text{dim},i'_\text{dep},d'_1,\ldots,d'_{i_\text{dep}}+ i_\text{dim},\ldots,d'_{|D|}) ) \\
   = \delta_{i_\text{dim},i'_\text{dim}} \cdot \delta_{i_\text{dep},i'_\text{dep}} \cdot \delta_{d_1,d'_1} \cdots \delta_{d_{|D|},d'_{|D|}} \cdot \delta_{d_{i_\text{dep}},0},
\end{align*}
where $\delta_{i,j} = 0$ if $i \neq j$ and $1$ if $i = j$ is the Kronecker delta and $n$ is a value that depends on the concrete state\footnote{We omit the indices indicating this dependency for readability.}.
Concretely, we distingish between two cases.
If there is a $k$ with $k > i_\text{dep}$ such that $d_k = 0$, then we choose the first (minimal) such $k$ and set $n = k - i_\text{dep}$.
If there is none, i.e. $d_i > 0$ for all $i > i_\text{dep}$, then we start back from $0$ and choose the smallest $k$ with $d_k = 0$ (without any additional requirements on $k$).
In this case we also set $n = k - i_\text{dep}$.
This means that, as their names suggest, \texttt{next\_dim} increases the dimension and \texttt{select\_dep} adds the current dependency to the set of strong dependencies if it was not previously added.
The action \texttt{next\_dep} increases the current available dependency, skipping those that have already been selected.
Accepting states are all state of the form $(i,j,d_1,\ldots,d_{|D|}), i,j,d_1,\ldots,d_{|D|} > 0$, where all dependencies have been chosen to be strongly carried.
Finally, the initial state is $(1,1,0,\ldots,0) \in S_\text{cons}$, starting with no dependencies selected.
We will discuss the rewards later, in Section~\ref{sec:rl_rewards}.
 
Note that while the state space depends on the concrete instance, i.e. the \ac{SCoP} being considered, the set of actions does not.
This is important since it allows us to learn a policy to navigate these spaces in a fashion that is independent of the problem instance.

\begin{figure}
	\centering
  \resizebox{0.45\textwidth}{!}{
    \includegraphics{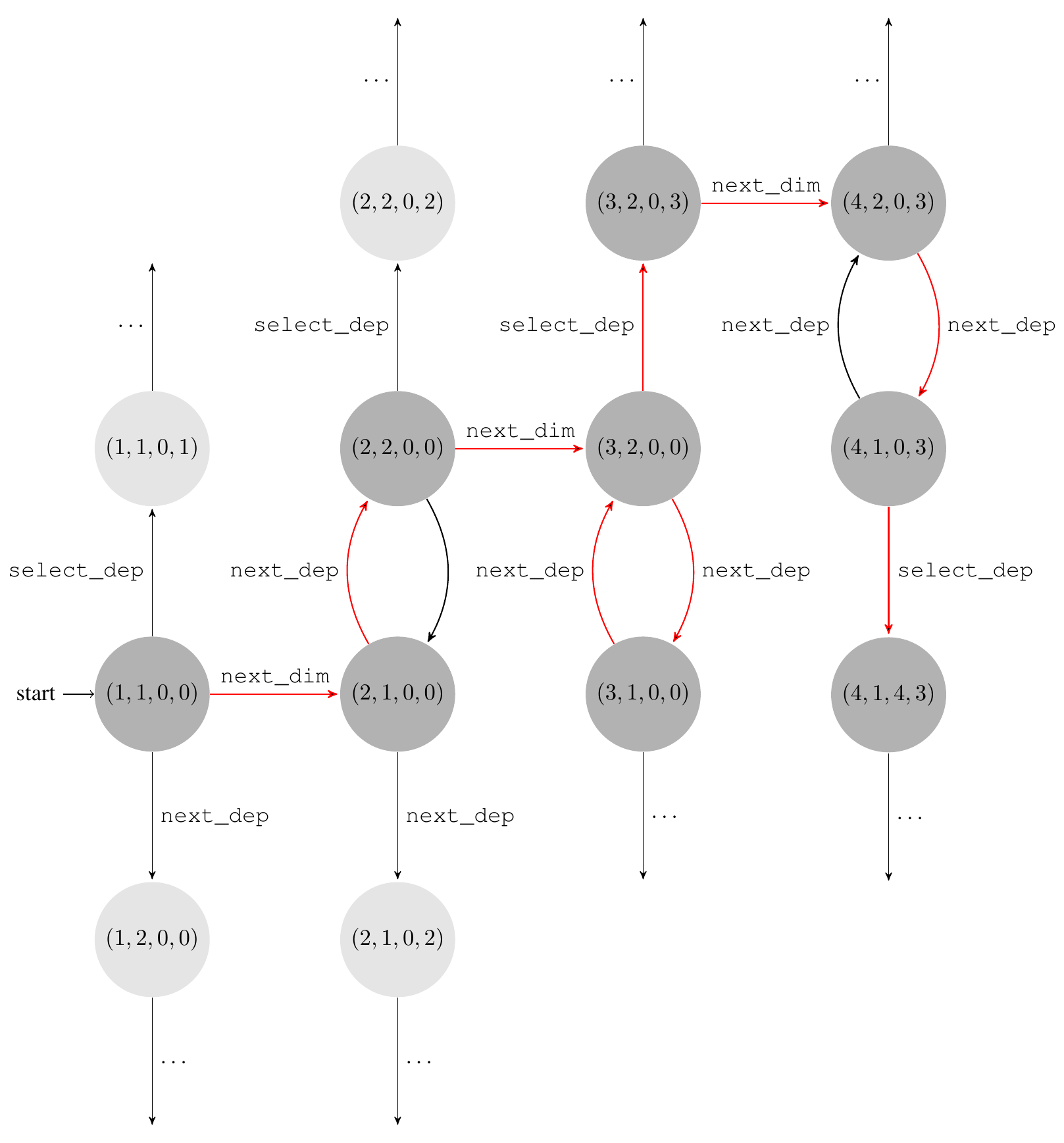}
  }
  \caption{An example of the schedule space construction \ac{MDP} and a concrete action sequence for the matrix-vector multiplication example.}
	\label{fig:construction_matvect}
\end{figure}
 
Consider the example in Figure~\ref{fig:construction_matvect}.
It shows the state space for the schedule space construction of the example of the matrix-vector multiplication from Section~\ref{sec:poly}.
The figure also shows an example run through the state space, corresponding to the sequence of actions
\texttt{next\_dim}, \texttt{next\_dep}, \texttt{next\_dim}, \texttt{next\_dep}, \texttt{next\_dep}, \texttt{select\_dep}, \texttt{next\_dim}, \texttt{next\_dep}, \texttt{select\_dep}.
The first action, \texttt{next\_dim}, increases the dimension, which is the first component $(1,\ldots) \rightarrow (2,\ldots)$.
Then, the \texttt{next\_dep} action changes the second index, indicating the current dependency: $(2,1,0,0) \rightarrow (2,2,0,0)$.
Since no dependency has been selected so far, the value of $j$ is $2$, corresponding to the the first case outlined above, yielding $n = 1$.
After increasing the dimension with a \texttt{next\_dim} action, the next \texttt{next\_dep} action again changes the dependency.
This time there is no dependency with index $>2$, and thus $j = 1$, by the second case, corresponding to $n = -1$.
A second \texttt{next\_dep} returns the second index, marking the dependency, to $2$, returning to the state $(3,2,0,0)$.
At this point the \texttt{select\_dep} adds an index to the current dependency ($2$), marking the current dimension to carry this dependency strongly: $(3,2,0,3)$.
The dependencies $S \rightarrow T$ and $T \rightarrow T$ are sorted, in that order.
This process continues until it reaches the state $(4,1,4,3)$.
This corresponds to a $4$-dimensional space, as indicated by the first component.
As indicated by the third component\footnote{Since the first two indicate the current dimension and dependency.}, the third dimension strongly carries the first dependency, $S \rightarrow T$, and the fourth dimension strongly carries the second dependency $T \rightarrow T$, as indicated by the last component. In particular, the accepting state itself uniquely defines the schedule space, as the order in which the dependencies are selected within a dimension is irrelevant to the construction, it only matters which dimensions specifically carry the dependencies strongly.
  
\subsection{Schedule space exploration}
Once the schedule space is generated, we look into how to find profitable schedules in this space.  
From Algorithm~\ref{algo:construction}, the schedule space is given by a vector $S = (G_1,G_2,\ldots,G_k)$, where each $G_d$ corresponds to the generators of a lattice polytope for the $d$-th dimension. 
A valid schedule consists precisely of a point in each of the lattice polytopes corresponding to the dimensions.
To construct each of these points we need to consider the generators $G_d$.
These are given as a set of vertices $v_1,\ldots,v_s$ and rays $r_1,\ldots,r_t$ as explained in Section~\ref{sec:poly}.
An arbitrary point $p$ is in the polytope iff it can be written as a convex combination of vertices and a positive linear combination of rays, i.e.
\[p = \sum_{i=1}^s \lambda_i v_i + \sum_{i=1}^t \alpha_i r_i,\]
where $\lambda_i \geq 0$ for all $i = 1,\ldots, s$ and $\sum_{i=1}^s \lambda_i = 1$ and $\alpha_i \geq 0$ for all $i = 1,\ldots,t$. Note that some work uses a third generator type, lines. This can always be converted to an equivalent set of generators as we define here.
We choose to have only rays instead of rays and lines as it is more uniform this way, making this representation more ameanable for \ac{RL}.
Schedules correspond to the points in the lattice polytope, which is a subset of this general polytope.
This means that $p$ has to have integer coefficients.

To formulate this as an \ac{MDP}, we introduce Algorithm~\ref{algo:exploration} which generates points in the polytopes following the same principles as~\cite{feautrier1992schedulingI,feautrier1992schedulingII,pouchet2007iterative,pouchet2008iterative,ganser2017iterative,ganser2018speeding}.
Once again, our algorithm leaves an unspecified, free function \texttt{select\_coeff}.
The goal of this function is to find the values of the coefficients $\lambda_i, i=1,\ldots,s$ and $\alpha_i, i=1,\ldots,t$.
We do this by iterating over all vertices and rays and selecting a coefficient in this iteration.
Like the authors in~\cite{ganser2017iterative}, we use a correction step, multiplying by the \ac{LCD} in the end to ensure the point is in the lattice polytope, i.e. has integer coefficients.
While this can be avoided through the design of the \texttt{select\_coeff} function, this correction step can simplify the design of the function.

\begin{algorithm}
	\caption{General exploration of the schedule space}
	\label{algo:exploration}
	\begin{algorithmic}[1]
	  \Input{A schedule space $S = (G_1,G_2,\ldots,G_k)$ as a vector, where each $G_i$ corresponds to the generators of a lattice polytope for the $i$-th dimension.}
	  \Output{A point in the schedule polytope for each dimension $p = (p_1,\ldots,p_k)$}
	  \For{ $i \in \{1,\ldots,k\}$}
    \State $v_1,\ldots,v_s,r_1,\ldots,r_t \leftarrow G_i$
    \State $p_i \leftarrow 0$
	  \For{ $x \in \{v_1,\ldots,v_s,r_1,\ldots,r_t\}$}
    \State $p_i \leftarrow p_i + \texttt{select\_coeff}(x) \cdot x$
    \EndFor
    \If{ $p_i \in \mathbb{Q}^n \setminus \mathbb{Z}^n$ }
    \State $p_i \leftarrow \operatorname{LCD}(p_i) \cdot p_i$
    \EndIf
    \EndFor
	  \Return $(p_1,\ldots,p_k)$
  \end{algorithmic}
\end{algorithm}

To define a corresponding \ac{MDP}, we start by defining the state space. For an integer $N > 0$, we define the space as the finite set\footnote{This is thus parameterized by $N$. Parameters like these are sometimes also called hyper-parameters, especially in the context of machine learning.} of the coefficients
\begin{align*}
  S_\text{expl}  = \{ (\lambda_{1,1},\ldots,\lambda_{1,s_1},\alpha_{1,1},\alpha_{1,t_1},\ldots,\lambda_{k,1},\ldots,\alpha_{k,t_k}) \\
  \mid \lambda_{i,j}, \alpha_{i,j} \in \{ 0, \ldots, N, \bot \} \text{ for all } i,j \}
\end{align*}
Note that since we have multiple polytopes $P_1,\ldots,P_k$, corresponding to the multiple dimensions, we use two indices for the generators, where the first index $d$ corresponds to the polytope $P_d$ and the second index iterates over the generators in $G_d$. This can also be conceptually understood as unrolling the two loops in Algorithm~\ref{algo:exploration} in the state space.
We use the symbol $\bot$ to mark coefficients that have not been selected, as this is distinct from selecting $0$ as a coefficient.
Choosing the coefficients $\lambda_i \in \{0, \ldots, N\}$ means that in most cases, $\sum_{i=1}^s \lambda_i > 1$.
In that case we norm the coefficients by building the convex combination as $\sum_{i=1}^s \lambda_i v_i / \sum_{i=1}^s \lambda_i$.
Additionally, when $s = 1$ it is clear that $\lambda_1 = 1$, since for a single point there is only one possible convex combination.
We thus remove the corresponding coefficients entirely from the state space\footnote{This was the case for almost all examples we evaluated in this paper.}.

Since we need to choose a coefficient for each term, we do not include actions to steer the exploration as we did for the schedule space construction.
Thus, the action space corresponds directly to the function \texttt{select\_coeff}.
We define actions \texttt{select\_coeff0}, \ldots, \texttt{select\_coeffN} accordingly.
We set
\begin{align*}
  \textbf{P}_\text{expl}((a_1,\ldots,\ldots,a_i,\bot,\ldots,\bot),\texttt{select\_coeffX}, \\
  (a_1,\ldots,\ldots,a_i,X,\bot,\ldots,\bot)) = 1,
\end{align*}

and for all other states $s,s' \in S_\text{expl}$ and actions $a \in \Act_\text{expl}$, we set $\textbf{P}_\text{expl}(s, a, s') = 0$.
The initial state corresponds the starting configuration with no coefficients defined, i.e., $(\bot,\ldots,\bot)$.

Similar to the space construction, a crucial aspect of this formulation is that the actions are independent of the \ac{SCoP}.
This way, a heuristic model can learn to find good schedules by selecting the correct coefficients.

Consider again the example of the matrix-vector multiplication kernel. 
With the schedule space constructed as depicted in Figure~\ref{fig:construction_matvect}, the schedule space has four dimensions, where the third carries the dependency $S \rightarrow T$ strongly and the fourth one $T \rightarrow T$.
This yields four polytopes, one for each dimension, which after using Chernikova's algorithm can be generated each by a single vertex and $10,10,10,13$ rays respectively. 
These all live in a $7$ dimensional vector space.
Note that neither the $4$ dimensions of the schedule space nor the $7$ dimensions of the polytopes for each schedule space dimension have a direct interpretation in terms of the loop bounds.
They represent loop schedules according to Farkas' lemma (cf. Section~\ref{sec:poly}) and they are not intuitively easy to understand in terms of the loops' \ac{AST}.
As mentioned above, the coefficient selection for the vertex is not part of the coefficient selection in the state space of the \ac{MDP}, since it is a single vertex in all four dimensions.

\begin{figure}
	\centering
  \resizebox{0.45\textwidth}{!}{
    \includegraphics{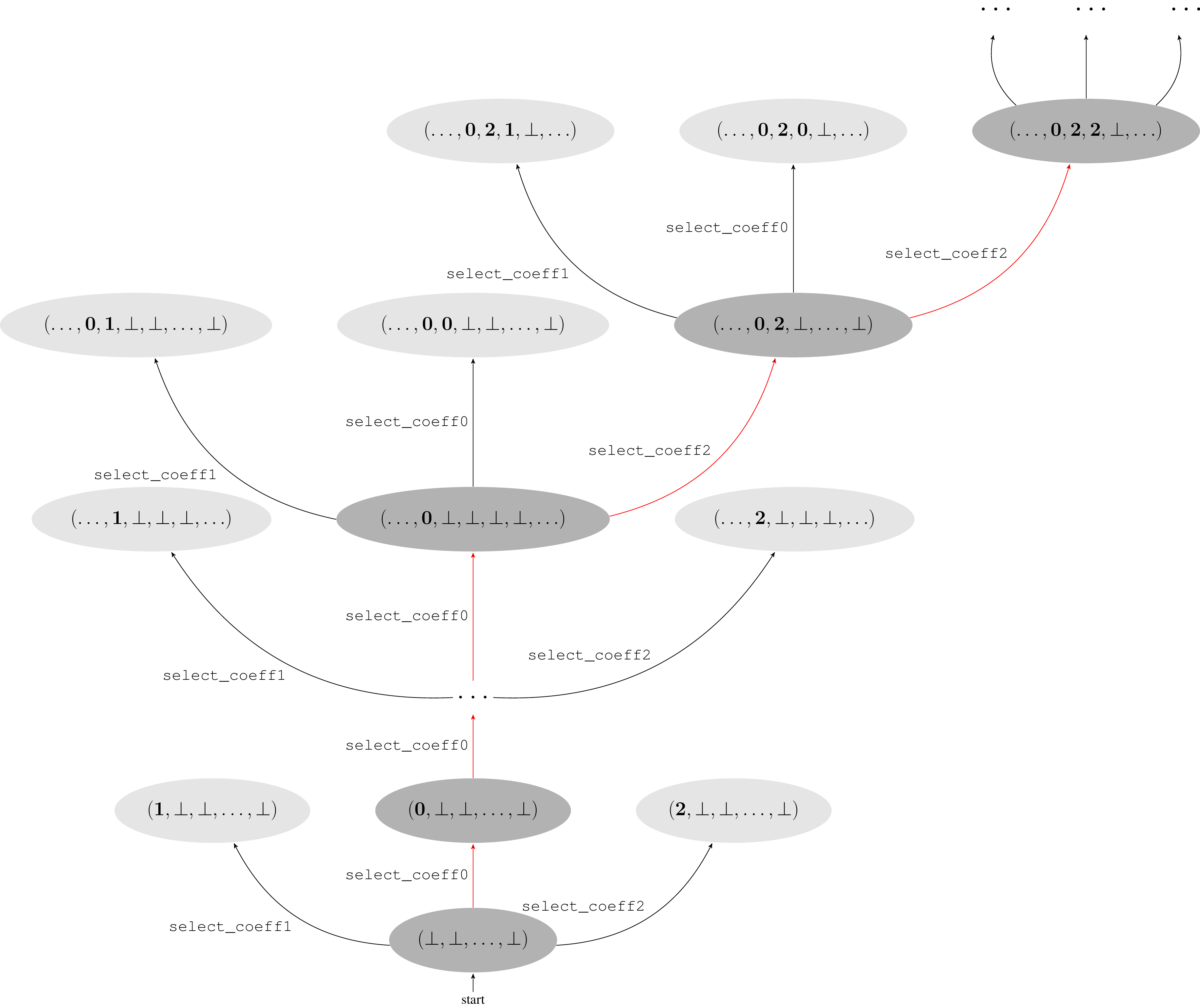}
  }
  \caption{An example of the schedule space exploration \ac{MDP} and a concrete action sequence for the matrix-vector multiplication example.}
	\label{fig:exploration_matvect}
\end{figure}
  
Figure~\ref{fig:exploration_matvect} shows the beginning of a sample run through this space, given by the following action sequence (omitting the \texttt{select\_coeff} part of the name):
\texttt{0}, \texttt{0}, \texttt{0}, \texttt{0}, \texttt{0}, \texttt{2}, \texttt{2}, \texttt{1}, \texttt{0}, \texttt{2}, \texttt{0}, \texttt{0}, \texttt{0}, \texttt{0}, \texttt{2}, \texttt{0}, \texttt{0}, \texttt{1}, \texttt{0}, \texttt{2}, \texttt{0}, \texttt{0}, \texttt{0}, \texttt{1}, \texttt{0}, \texttt{2}, \texttt{0}, \texttt{0}, \texttt{0}, \texttt{0}, \texttt{0}, \texttt{0}, \texttt{2}, \texttt{0}, \texttt{0}, \texttt{0}, \texttt{0}, \texttt{0}, \texttt{0}, \texttt{0}, \texttt{0}, \texttt{0}, \texttt{0}.
The coefficients of the rays are selected dimension by dimension: $\alpha_{1,1} = 0, \ldots, \alpha_{1,6} = 2, \alpha_{1,7} = 2, \alpha_{1,8} = 1, \ldots,\alpha_{1,10} = 2,\alpha_{2,1} = 0,\ldots,\alpha_{4,13} = 0$.
After selecting all coefficiens with that sequence, the resulting points are given by:
\begin{align*}
  p_1 =~ & (0,0,0,0,0,0,0)~ + (0,0,0,0,0,-1,0) \\
     & + 2 \cdot (0,0,0,0,0-1,-1)~ + 2 \cdot (0,0,0,-1,-1,0,0) \\
  &  + (-1,-1,0,0,0,0,0)~  = (-1,-1,0,-2,-2,-3,-1) \\
 \vdots \\
  p_4 =~&  (0,0,1,0,0,0,0)~ + 2 \cdot (0,0,1,0,0,0,0) \\
  &    = (0,0,3,0,0,0,0).
\end{align*}

We see how this \ac{MDP} can produce concrete schedules.
Finally, this points can be translated into a valid transformation $(i,j) \rightarrow (k,l)$ as explained in Section~\ref{sec:poly}.
This translates to the following transformed C kernel:
\begin{minted}{C}
if (N >= 1)
  for (int k = -N; k <= 0; k += 1) {
    if (N + k >= 1)
      for (int l = 0; l < N; l += 1)
T:      y[-k] += y[-k][l]*x[l];
    if (k <= -1)
S:    y[-k - 1] = 0;}
\end{minted}

Note again that the correspondence between the representation as points in the schedule space and the transformed kernel is not directly visible.
The dimensions of a multidimensional schedule define how the individual points in the schedule are ordered (lexicographically), it does not directly translate e.g. to loop nesting.
\todo{See if I can make the example very explicit.}
      
This concrete example was in fact found by a simple heuristic, with a bias towards defining coefficients to be $0$. 
We describe the heuristic in Section~\ref{sec:eval}. 

\subsection{Rewards}
\label{sec:rl_rewards}
 
The rewards of an \ac{MDP} define a feedback loop -- the quality of an action taken in a given state.
This is needed for a learning algorithm so it can decide about what action is favorable to take in a given state and adjust the agent's model in case the action was infavorable.
Reinforcement Learning can then be used to train an agent that navigates an \ac{MDP}.
Recent work has shown that it is possible to use Deep Neural Networks as agent models that are able to generalize over the large search space and able to learn to make good decisions on in previously unseen states~\cite{mnih2015human,silver2016mastering}.

\begin{figure}
	\centering
    \tikzsetnextfilename{ext_pipeline}
    \input{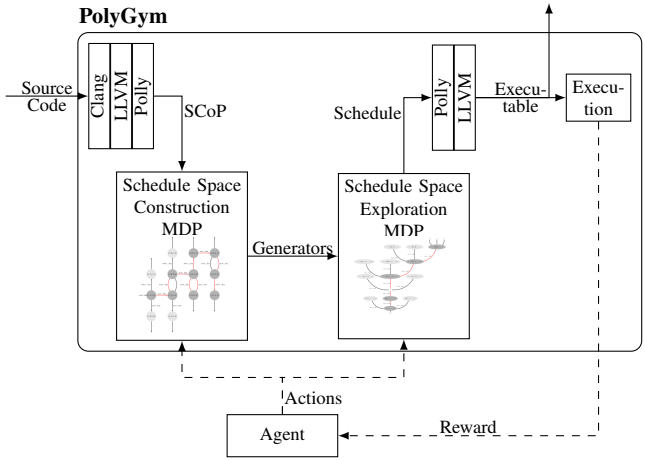}
  \caption{The PolyGym flow}
	\label{fig:pipeline}
\end{figure}

Figure~\ref{fig:pipeline} shows the flow with two \ac{MDP} models in PolyGym in a \ac{RL} context.
Note that PolyGym does not define an agent, it only defines the environment to learn with such an agent.
The two separate \acp{MDP} produce a single schedule, which is then compiled with LLVM/Poly and executed, to determine the reward.
There is no immediate reward to the construction of the schedule space, nor to the partial states in the exploration step, since these do not define an actual schedule. 
As such, we define the rewards for both individual \acp{MDP} uniformly, based on the final constructed schedule, as follows.
If the action leads to a complete schedule, which is a terminal state, then we give the speedup over a traditional optimizing compiler, e.g., using LLVM with the \texttt{-O3} flag,  as a sample-independent metric as reward.
If it leads to an incomplete schedule, we give $0$ as reward.
Finally, if it leads to any invalid state, we give negative reward. Invalid states can occur when constructing an empty polytope in the \ac{MDP} of the schedule space construction.
    They can also occur in the exploration \ac{MDP}, when there are multiple vertices in a polytope and all vertex coefficients are selected to be $0$ (leading to no convex combination).
  
In this way, (positive) rewards are only given for the final construction which obtains a valid schedule.
This constitutes what is called a sparse reward setting.
  For example, the schedule for the matrix-vector multiplication kernel\footnote{For the vector dimension $N=2000$, with data generated procedurally using the Polybench tools.} found by the examples described in Figures~\ref{fig:construction_matvect} and~\ref{fig:exploration_matvect}, the resulting kernel runs an impressive $140 \times$ faster than the original kernel compiled with \texttt{-O3}\footnote{The full setup is described in Section~\ref{sec:eval}.}.
This level of improvement is thus not easy to obtain in more complex kernels and larger input sizes.
 Thus, the final action \texttt{select\_coeff0} (not shown in Figure~\ref{fig:exploration_matvect} for space reasons) has a reward of $140$, while all other actions have a reward of $0$.
State-of-the-art methods in \ac{RL} trickle down this reward to the other actions~\cite{mnih2015human,silver2016mastering}.
Importantly, this reward should also be trickled down to the schedule space construction in a learning setting, even if the \ac{MDP} is conceptually a different one.

\subsection{Limitations}

While our formulation of the schedule search problem as an \ac{MDP} with an instance-independent action set enables reinforcement learning for this problem, it still has some limitations.

An important limitation is that some accepting states do not yield schedules, as discussed previously.
There are two possobilities of this happening. 
The first corresponds to the schedule space construction.
Not every schedule space constructed with Algorithm~\ref{algo:construction} actually has schedules.
If too many dependencies are carried strongly in a dimension, the intersection of the corresponding polytopes can be empty, leading to a space without any valid schedules.
The randomized construction of~\cite{ganser2017iterative}, similarly, does not provide such a (formal) guarantee.
In practice, however, biasing the construction towards many dimensions, as they do, yields non-empty spaces.
We are confident this fact can easily be learned by pertinent methods.

The second possibility has to do with the vertices in a generating set for a polytope.
If all vertices have a coefficient of $0$, Algorithm~\ref{algo:exploration} does not produce a schedule, since the point constructed is not in the lattice polytope.
The authors in~\cite{ganser2017iterative} choose the vertex at random.
We cannot do the same, as it breaks the \ac{MDP} abstraction, where the reward of a state is non-deterministic.\footnote{Since the execution times of the \acp{SCoP} are non-deterministic as well, this is the case for all rewards. However, if the statistical variance of execution times is not negligible, then the whole problem of selecting an optimal schedule is ill-posed in the first place. This is not the case in practice.} 
We did not run into this problem in any example considered in this paper.
Having more than one vertex seems to be an extremely rare occurrence in practice. 
This is both a benefit and a problem: While we mostly do not have to deal with this problem, if at some point we ever do, we will probably not have enough samples to learn to deal with it.

A final limitation corresponds to the exploration phase.
There are two ways in which we exclude some points.
By choosing a finite $N$ as a hyper-parameter, we technically limit the possible coefficients of the schedule space.
This is necessary, however, since otherwise there is an infinite amount of possible schedule polytopes~\footnote{Many of these are equivalent, since the number of orderings is finite}.
Similarly, the fact that we use integer coefficients and calculate the \ac{LCD} in Algorithm~\ref{algo:exploration} might exclude some points that would otherwise be found with rational coefficients. 
With an unbounded $N$ this problem would not exist.
Consequently, a larger $N$ mitigates it.

While the limitations discussed here technically exclude some solutions or allow algorithms~\ref{algo:construction} and~\ref{algo:exploration} to fail, they do so only in rare corner-cases.
Moreover, our reward design and the hyper-parameter $N$ allow us to avoid or at least mitigate these limitations in those rare corner-cases.

\section{Evaluation}
\label{sec:eval}
\begin{figure*}[h]
	\centering
  \resizebox{\textwidth}{!}{
    \includegraphics{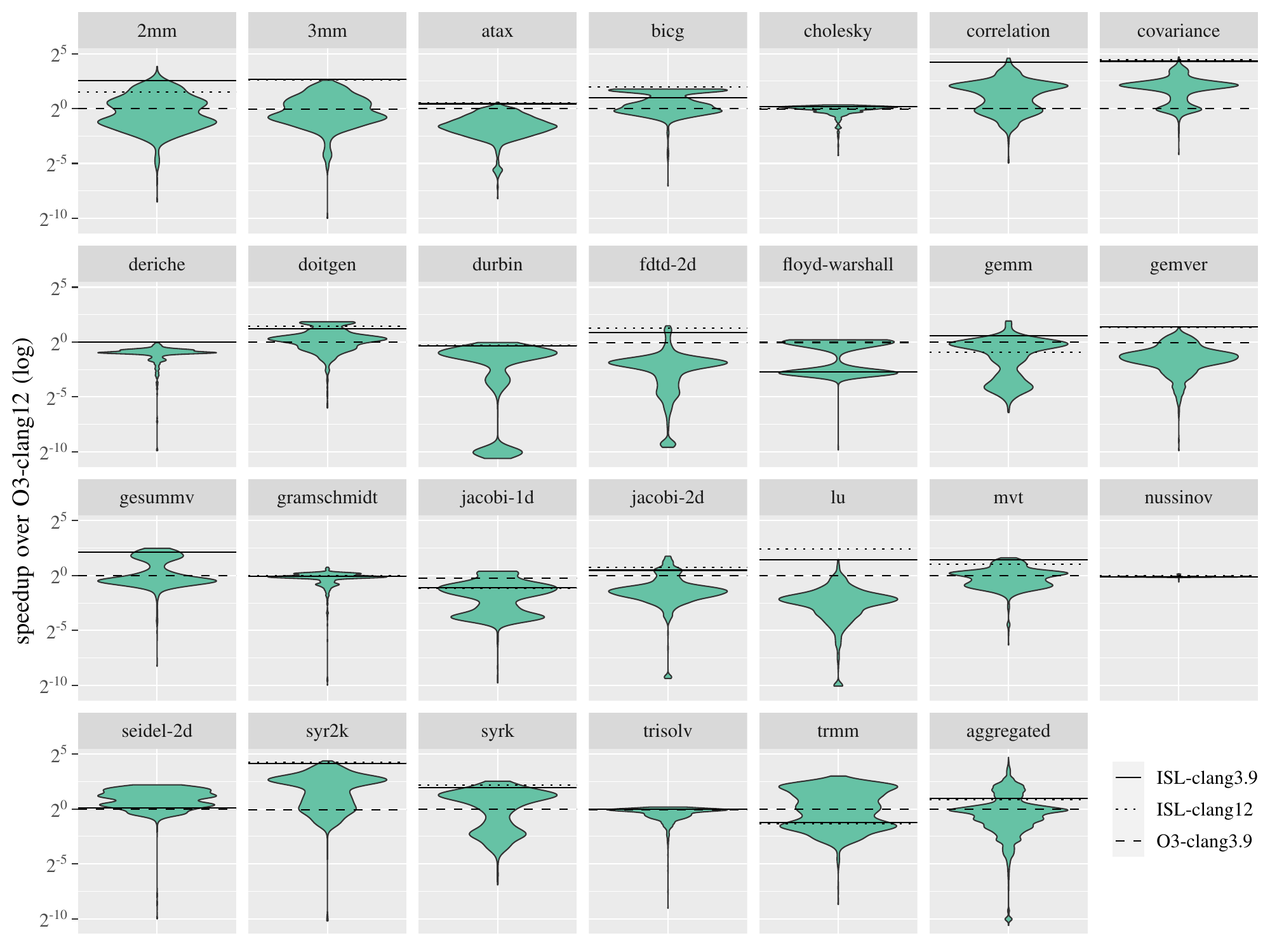}
  }
  \caption{Violin plot of the distribution of speedups by benchmark, with speedups as reference lines (geometric mean, 4 executions).}
  \label{fig:speedup_distributions}
\end{figure*}
   
\begin{figure*}[h]
	\centering
  \input{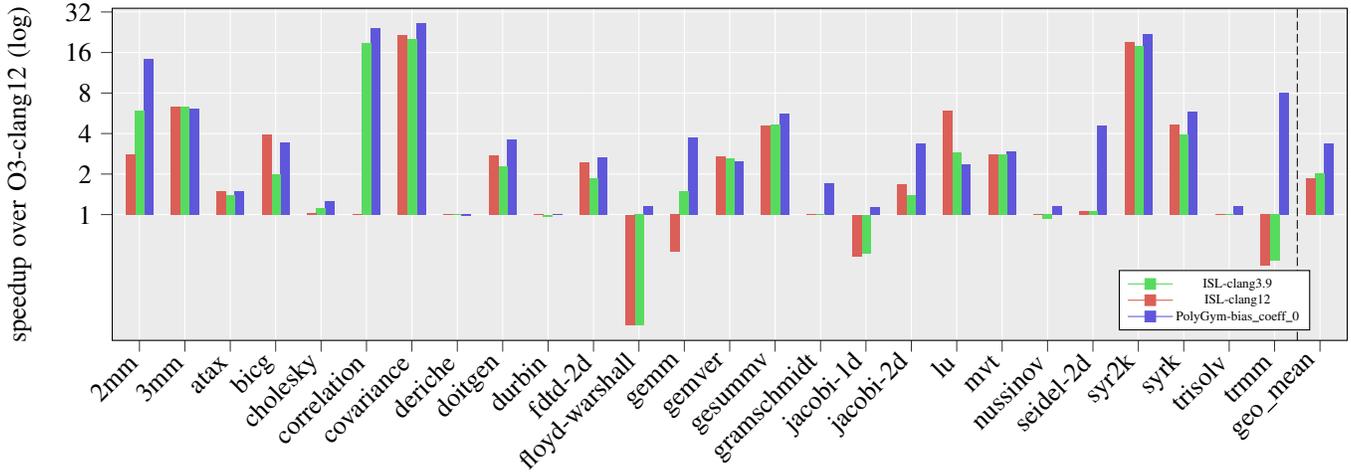}
  \caption{Speedups over Clang O3 of individual kernels of Polybench. PolyGym was configured with the \emph{select\_coeff\_0} heuristic and 1000 iterations.}
  \label{fig:speedups_by_sample}
\end{figure*}
 
In this section we evaluate the performance potential of the search space of our \ac{MDP} formulation by exploring it with simple heuristics.
We compare the best schedules we find in a fixed number of explorations with the ones from the state-of-the-art heuristic implemented in ISL and Polly in recent versions.
In the following, we describe the experimental setup we used and discuss the results obtained.

\subsection{Experimental setup}
We implemented the \acp{MDP} of the schedule space as an OpenAI Gym Environment, so it is conveniently usable for future machine learning heuristics~\cite{brockman2016openai}.
The algorithms are implemented in Python, relying on the ISL binding islpy\footnote{\url{https://github.com/inducer/islpy}} for operations on integer sets.
To transform the polytopes into the generator representation, we used Polyite's implementation of the Chernikova algorithm\footnote{\url{https://github.com/stganser/chernikova}}.
Finally, we use LLVM and Polly in version 3.9 to extract polyhedral representations of SCoPs and to transform them according to the computed schedules.

All schedules were evaluated on the same system, consisting of an AMD 3960X with 24 cores and 48 threads, organized in 8 units, each with 3 cores. 
The units have a dedicated 16 MB L3 cache. 
Each core has a dedicated 32 KB L1 and a 512 KB L2 cache.
The benchmarks are executed in a single unit, that is, using at most 3 cores (and 6 threads).
We do so because of the low interference between units, which allows us to run up to four experiments in parallel without adding too much noise in the observations. The configuration of four units has been identified to have necligible side-effects in preliminary experiments.

For our evaluation we use the Polybench benchmark suite~\cite{polybench}, which consists of $30$ numerical computations from various domains, such as image processing, statistics, and linear algebra.
Each of the benchmarks contains a \ac{SCoP} kernel, i.e. a loop nest that satisfies the conditions of the polyhedral model.
We had to exclude the \texttt{ludcmp}, \texttt{heat-3d}, and \texttt{adi} kernels because of their large number of dependence polytopes.
The sheer size of these polytopes significantly slows down the runtime of the Chernikova algorithm and thus the construction of the schedule space.
 Since programs generated by unfavorable schedules could potentially run for a long time, we set a timeout that is proportional to the LLVM O3 runtime by a factor of 10 when executing.

We compare the schedules found by PolyGym against standard optimizing compilers and against ISL. 
As standard optimizing compiler we use LLVM in version 12 with the \texttt{-O3} flag (LLVM O3).  The ISL experiments were run with flags \texttt{-polly -polly-parallel=true -polly-vectorizer=none -polly-tiling=true -polly-default-tile-size=64}.
In the experiments, we used 4 measurements and report the minimum to eliminate measurement inaccuracies.

\subsection{Result analysis}

To analyze how many profitable schedules the space spawned by our MDP formulation includes, we generate $1000$ schedules per benchmark kernel with the a simple bias the exploration of the schedule space towards the \texttt{select\_coeff0} action. This bias results in overall less complex schedules.
Figure~\ref{fig:speedup_distributions} shows the distributions of speedups by individual samples, along with the performance of ISL and LLVM O3.
We can observe that the search space contains many profitable schedules.
Many benchmarks, like \texttt{2mm},\texttt{doitgen}, \texttt{gemm},\texttt{jacobi-2d} or \texttt{seidel-2d}, show a sigificant distribution of points better than those found by ISL.
This suggests that an agent could learn to achieve this better performance results without iteratively executing the kernel.

Figure~\ref{fig:speedups_by_sample} shows the maximum measured speedups for individual kernels of the polybench suite.
We find schedules with an overall speedup of $3.39$x over O3-clang12, which is $1.83$x faster than the ones of ISL-clang12 and $1.67$x faster than ISL-clang3.9.
For 20 of the 26 kernels, the heuristic iteratively finds more profitable schedules than ISL-clang12, for 22 of 26 for ISL-clang3.9.
The results are not directly comparable to Polyite~\cite{ganser2017iterative,ganser2018speeding}, because they we use a different hardware system. However, the overall results seem to be comparable in terms of the improvement, which is not surprising, since we use a similar search space and a similar random sampling process.
Compared to Polyite, our MDP formulation is shape-agnostic.
This can enable an agent to learn to navigate this space without requiring an iterative execution.

\begin{figure}[h]
	\centering
  \input{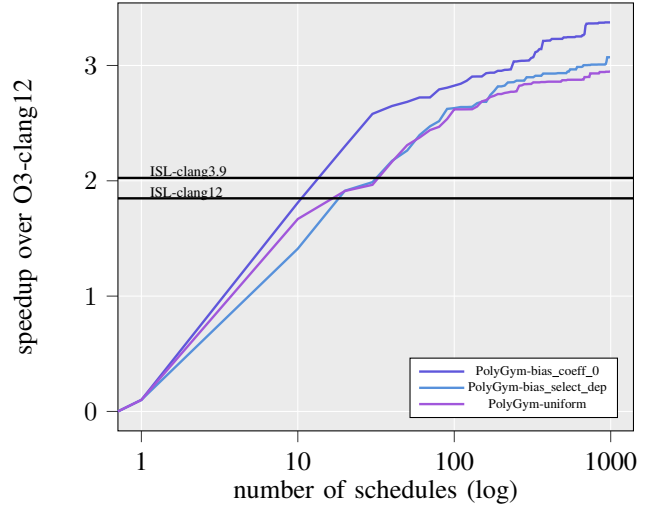}
  \caption{Maximum speedup over LLVM O3 achieved by the different heuristics for action selection with increasing number of explored schedules. 
    The plot depicts the geometric mean over all selected benchmarks of the Polybench suite. 
    As reference, the aggregated speedups of ISL over LLVM O3 in different version (ISL-clang3.9, ISL-clang12) is also included.}
  \label{fig:speedup_development_by_heuristics}
\end{figure}
 
We further analyze the influence that a potential heuristic has on the sampling process.
We employ different trivial heuristics for demonstration.
\begin{itemize}
  \item In \emph{bias\_select\_dep}, we bias the schedule space construction phase towards the \texttt{select\_dep} action, which results in less schedule dimensions.
  \item In \emph{bias\_coeff\_0}, we bias towards the \texttt{select\_coeff0} action in the schedule space exploration phase.
  \item In \emph{uniform}, we select actions uniformly at random.
\end{itemize}

Figure~\ref{fig:speedup_development_by_heuristics} shows the performance of the PolyGym search space with different heuristics on the Polybench suite.
The plot shows the geometric mean across the different benchmarks as the number of schedules sampled increases.
We see that different heuristics impact the space exploration achieving different performances overall.
All evaluated heuristics outperform the schedules found by ISL-clang3.9 and ISL-clang12 in less than $40$ sampling iterations.
Most notably, when biasing the selection towards the coefficient 0 (\emph{bias\_coeff\_0}), the cross-over is at 11 iterations for ISL-clang12 and after 13 iterations for ISL-clang3.9.
Crucially, this shows that there is potential to learn to explore this space efficiently.
All of the evaluated heuristics, however, are very simple and make not use of the MDP's states.
In future work, this heuristic can be a Deep Neural Network that is optimized to take the best actions using Reinforcement Learning algorithms.

\section{Conclusions}
\label{sec:conclusions}
In this paper we presented PolyGym -- an environment for polyhedral schedule optimizations for \acf{RL}, based on a generic, SCoP-independent \acf{MDP}.
With this formalization, it is possible to develop heuristic with RL algorithms that learn models that produce profitable schedules directly, without iterative benchmarking.
Using the PolyBench suite, we have shown that the search space offers the potential to learn to produce results that significantly outperform state-of-the-art heuristics.
PolyGym is usable as an environment conforming to the Gym interface, which is widely-used by RL algorithms, enabling its integration with minimal effort.
Therefore, PolyGym allows to investigate learning models in this task without diving into the theory of the polyhedral models, which is a high barrier.
We believe that this will expose polyhedral optimizations as a learning problem, advancing the field of machine learning for compilers.

In future work, we plan to investigate models and \ac{RL} algorithms to learn profitable schedules.
We also plan to integrate reward functions for further objectives, such as energy consumption, into PolyGym.

\section*{Acknowledgments}
This work was funded in part by the German Federal Ministry of Education and Research (BMBF) within the project ScaDS AI (BMBF 01IS18026B), the German Research Council (DFG) through the TraceSymm project (number 366764507) and the Studienstiftung des deutschen Volkes.

\bibliographystyle{IEEEtran}
\bibliography{references}

\end{document}